# Programming by Examples Meets Historical Linguistics: A Large Language Model Based Approach to Sound Law Induction

**Atharva Naik**[1] **Darsh Agrawal**[1] **Hong Sng**[1] **Clayton Marr**[3] **Kexun Zhang**[1] **Nathaniel Robinson**[2]
**Kalvin Chang**[1] **Rebecca Byrnes**[1] **Aravind Mysore**[1] **Carolyn Rose**[1] **David Mortensen**[1]
Carnegie Mellon University[1] John Hopkins University[2] Ohio State University[3]
{arnaik,darsha,hongsng,kexunz,rbyrnes,amysore2,cprose,dmortens}@cs.cmu.edu
nrobin38@jhu.edu kalvinc@alumni.cmu.edu

## Abstract

Historical linguists have long written "programs" that convert reconstructed words in an ancestor language into their attested descendants via ordered string rewrite functions (called SOUND LAWS) However, writing these programs is error-prone, motivating the development of automated SOUND LAW INDUCTION (SLI) which we formulate as Programming by Examples (PBE) with Large Language Models (LLMs). While LLMs have been effective for code generation, recent work has shown that PBE with LLMs is challenging but improvable by fine-tuning, especially with training data drawn from the same distribution as evaluation data. In this paper, we create a conceptual framework regarding what constitutes a "similar distribution" for SLI and propose four kinds of synthetic data generation methods with varying amounts of inductive bias to investigate what leads to the best performance. Based on the results, we create a SOTA open source model for SLI as PBE (+6% pass rate with a third of the parameters of the next best open source LLM).

## 1 Introduction

In the 19th century, European linguists, then called "philologists," made a striking discovery: the sounds of the words of languages change in a principled, rule-governed way (Brugmann and Osthoff, 1878). When formulated, these rules are called SOUND LAWS. Linguists determined that the ancestral forms (protoforms) of words could be reconstructed by comparing their descendant words (reflexes) across a language family, a set of languages sharing a common ancestor. The words in any member of the language family could then be derived by applying an ordered sequence (cascade) of sound laws (Marr and Mortensen, 2020), to the protoforms in the reconstructed ancestor language. In writing these cascades of sound laws, Neogrammarian linguists were essentially writing programs—composed sequences of regular expression replacements like

$$\emptyset > \text{k}/__\left\{ \begin{array}{c} \text{i} \\ \text{u} \end{array} \right\}\#$$

("rewrite the empty string with /k/ after /i/ or /u/ at the end of a word", a sound change in the history of Huishu). In this paper, we investigate the capability of LLMs to write programs representing cascades of sound laws.

This task, where input-output pairs are given and a program mapping between input and output is generated, is called Programming by Examples (PBE) (Gulwani, 2016). While various computational models for Programming by Examples exist, LLMs—which have proved remarkably adept at code generation in general—have struggled with PBE (Li and Ellis, 2024; Fu et al., 2024). As Li and Ellis (2024) observed, this weakness can be mitigated if the LLMs are finetuned on data drawn from the same distribution as real-world data. But in what sense should they be in the same distribution? A concept from linguistics proves useful here (both for the linguistic and general cases): structure (form) versus substance. The debate about the relative importance of structure and substance goes back to the late 19th century, with linguists like Saussure (De Saussure, 1879), Hjelmslev (Hjelmslev, 1961), and Bloomfield (Bloomfield, 1926) advancing the structuralist view and Diver (Diver, 2012), Venneman (Vennemann and Ladefoged, 1971; Vennemann, 1972), Dressler (Dressler, 1984), Bybee (Hooper, 1979), Ohala (Ohala, 2017) and many other scholars advancing the substantive view (Boye and Engberg-Pedersen, 2015).

Sound laws that rewrite /t/ as /d/ at the end of a word and /d/ as /t/ at the end of the word are formally the same ($a > b/L_R$, $a$ is rewritten as $b$ between the left context $L$ and the right context $R$). However, d > t / _# is much more likely to be found in a cascade of actual sound laws than t > d

/ _ #. That is, from the standpoint of substance, d > t / _# should be a better finetuning example.

At the outset of our research, we hypothesized that finetuning examples that were most like test cases in substance would provide the greatest benefit. This hypothesis was partially vindicated—finetuning works best when the input side of the I/O pairs resembles a word in a human language. However, when we attempted to constrain sound laws so that they resembled actual laws in substance, we observed much worse performance than when the laws were structurally well-formed but substantively random (perhaps because randomness prevents the models from acquiring pernicious biases from too-uniform sets of actual rules).

Our contributions are as follows:

1. A Programming by Examples (PBE) formulation of sound law induction (SLI), leveraging advancements in program synthesis.
2. The creation of a dataset for SLI as PBE with Python code and benchmarking of GPT-4 (Achiam et al., 2023) and various open source LLMs for SLI as PBE.
3. An investigation into the impact of structure vs substance, for designing the distribution of inputs and programs for creating synthetic fine-tuning data to improve SLI as PBE.
4. PySLICoder, a SOTA open-source model for SLI trained using GPT-4 generated data with the best synthetic data generation algorithm identified in the study.

## 2 Conceptual Framework

Programming by Examples or PBE (Gulwani, 2016), is the task is of finding a program $\rho$ that maps each input $X$ to a corresponding $Y$ given input-output pairs $(X, Y)$ as $\rho(X) = Y$. Given this definition, one can consider Sound Law Induction or SLI as PBE by treating the protoforms from the ancestral language as the inputs $X \in \Sigma*$, the corresponding attested forms in the descendant language as the outputs $Y \in \Sigma*$ and the sound laws as the string rewrite programs $\rho$. Additionally, any data generation procedure can be categorized as sampling inputs $X$ from a distribution $\mathcal{P}_X$ and programs $\rho$ from a distribution $\mathcal{P}_\rho$ i.e. $X \sim \mathcal{P}_X, \rho \sim \mathcal{P}_\rho$. Given $(X, \rho)$, the $(X, Y)$ input-output examples can be constructed as $Y = \rho(X)$ by simply executing the programs on the inputs.

Recent work on PBE with Large Language Models (LLMs) (Li and Ellis, 2024) has shown that they struggle at this task, but can get better through fine-tuning. However, this work also shows that closer the distribution of the fine-tuning data (in other words the distribution of inputs $\mathcal{P}_X$ and programs $\mathcal{P}_\rho$) is to the evaluation data, the better the performance.

This could mean one of two things. Consider the case of a PBE model that extracts telephone numbers from email signatures. The fine-tuning data might consist of examples including strings similar to +99 999 999 99 99 or (999) 999-9999. In this case, the fine-tuning data would come from the distribution defined by **substantive** similarity to $\mathcal{P}_X$. However, the distribution could be defined in terms of **logical structure** (where the telephone number—the string extracted—would be the last line in the signature and its substance being, in the extreme case, random). Neither the substance-based distribution or the structure-based distribution is likely to correspond exactly to the distribution of telephone numbers in email signatures. Likewise, it is an empirical question whether the distribution of naturalistic proto-languages $\mathcal{P}_X$ and sound laws $\mathcal{P}_\rho$ will best align with substance-based or structure-based fine-tuning data. Indeed, substance and structure point to different ends of the spectrum of inductive biases required by the fine-tuning data. We consider the two ends to represent approaches purely focused on structure (low bias) and substance (high bias) on either end, to represent the debate about their relative importance (Boye and Engberg-Pedersen, 2015).

To investigate where on this spectrum the best performance can be achieved, we develop four types of synthetic data generation methods with progressively greater focus on substance in the input $\mathcal{P}_X$ or program distributions $\mathcal{P}_\rho$ as shown in Figure 2. On the structure end, we begin with a low-bias approach **RP-RI** which samples random string manipulation programs and random inputs where the programs apply. **RP-LI** adds more substance by generating “pronounceable” nonce word inputs and applicable programs by prompting an LLM. **RP-PI** adds more substance by giving the LLM randomly sampled words from Proto-Oceanic (poc) and Proto-Tangkhulic (ptk), two ancestral languages with words similar to the inputs in our SLI evaluation benchmark, as inputs. Finally **IDP-PI** has the most substance. It incorporates real sound laws from other languages in from the Index Diachronica database (Index Diachron-ica Contributors, 2016).

## 3 Related Work

### 3.1 Programming by Examples

Code generation from input-output examples that demonstrate the expected behavior is well explored in the literature of programming by examples (PBE) (Gulwani, 2016). Researchers in this space have explored ways of leveraging input-output examples to guide programming synthesis in neural models (Ye et al., 2021; Shi et al., 2023) with recent efforts focused on prompting LLMs with examples or test-case information for program synthesis (Austin et al., 2021; Lahiri et al., 2022; Jain et al., 2022). Austin et al. (2021) benchmark LLMs for program synthesis in Python from input-output examples and natural language (NL) intents, by proposing the Mostly Basic Programming Problems (MBPP) dataset. Lahiri et al. (2022) propose an interactive approach for formalizing underspecified intents by obtaining user feedback on proposed input-output examples. Recent work (Li and Ellis, 2024; Fu et al., 2024) has shown that LLMs are not effective at PBE by default but can be improved by fine-tuning as long as the train and test distribution overlap (Li and Ellis, 2024).

### 3.2 Automatic sound law induction

This topic area has seen a significant increase in activity as of late. Chang et al. (2023) apply Albright and Hayes (2003)'s deterministic string algorithm (Wilson and Li, 2021) to automate sound law induction. Sims-Williams (2018) proposes the task of computerized forward reconstruction (CFR), and Marr and Mortensen (2020) present a CFR system equipped with a diagnostic suite to assist cascade refinement. Luo (2021) proposes a Monte Carlo Tree Search (MCTS) algorithm for automatic derivation of sound laws, while List (2024) uses layers of functionally simultaneous sound changes. Our approach, meanwhile, performs both CFR and sound law induction with greater sample efficiency. More broadly, to our knowledge, we are the first to perform forward reconstruction using large language models.

### 3.3 Linguistic Reasoning with Large Language Models

LLMs have a mixed record when it comes to linguistic reasoning. ChatGPT underperformed in generalizing morphological patterns to nonce words relative to humans and trivial baselines (Weissweiler et al., 2023). GPT-3.5-turbo-instruct could analyze compositional word formation and derivation in German compounds, but could not identify illicit derivations (Weller-Di Marco and Fraser, 2024). Zhou et al. (2024) found that LLMs struggle at classifying so-that constructions but Mortensen et al. (2024) found that they demonstrate human-like lexical-syntactic flexibility. Whereas prior work focused on morphology, syntax, and semantics, we are the first to use LLMs for historical linguistics, generally, and historical phonology, specifically. In the realm of symbolic reasoning for linguistic pattern recognition (Vaduguru et al., 2021) have shown that PBE techniques can be leveraged for sample efficient linguistic generalizations using FlashMeta (Polozov and Gulwani, 2015) a data-driven PBE framework. Moreover, Luo (2021)'s MCTS algorithm for sound law induction essentially generates simple regular expression programs to represent sound laws. Motivated by these approaches and the observation that LLMs when coupled with fine-tuning can support PBE, we propose synthetic data generation and fine-tuning methods for sound law induction.

## 4 Method

Based on the spectrum of fine-tuning data introduced in section 2 we create four equally sized (2.5k instances) synthetic datasets with progressively increasing degrees of substantive bias and fine-tune Magicoder (Wei et al., 2024) on them using a specially designed prompt for SLI as PBE (Figure 3). We then evaluate each fine-tuned model on our SLI benchmark using the same prompt by generating $s(= 20)$ programs per instance as shown in Figure 1. The best condition determined is then used to generate more data and train our proposed model **PySLICoder** from Magicoder. We will now cover the SLI-as-PBE prompt, synthetic data algorithms, and fine-tuning details.

### 4.1 Sound Law Induction Prompting

To frame the task of SLI-as-PBE, LLMs are instructed to create sound law programs using a specific format, captured by a class called $\mathrm{BasicAction}$. The input prompt contains protoform/reflex pairs tokenized using PanPhon (Mortensen et al., 2016) as exemplified in Figure 3. The prompt contains six main components including instructions, code, and input-output examples. The first and second sections describe the task, evaluation measures, and the BasicAction class we use to represent sound laws as Python code. The third section represents the protoforms and reflexes and their tokenized versions. The fourth and fifth section contain code examples demonstrating usage of the BasicAction and the source definition for it. Finally, the sixth section contains additional instructions, such as to avoid importing other packages and to not modify or repeat the definition of BasicAction. The BasicAction class and its input specification are described in greater detail in Appendix B.1.

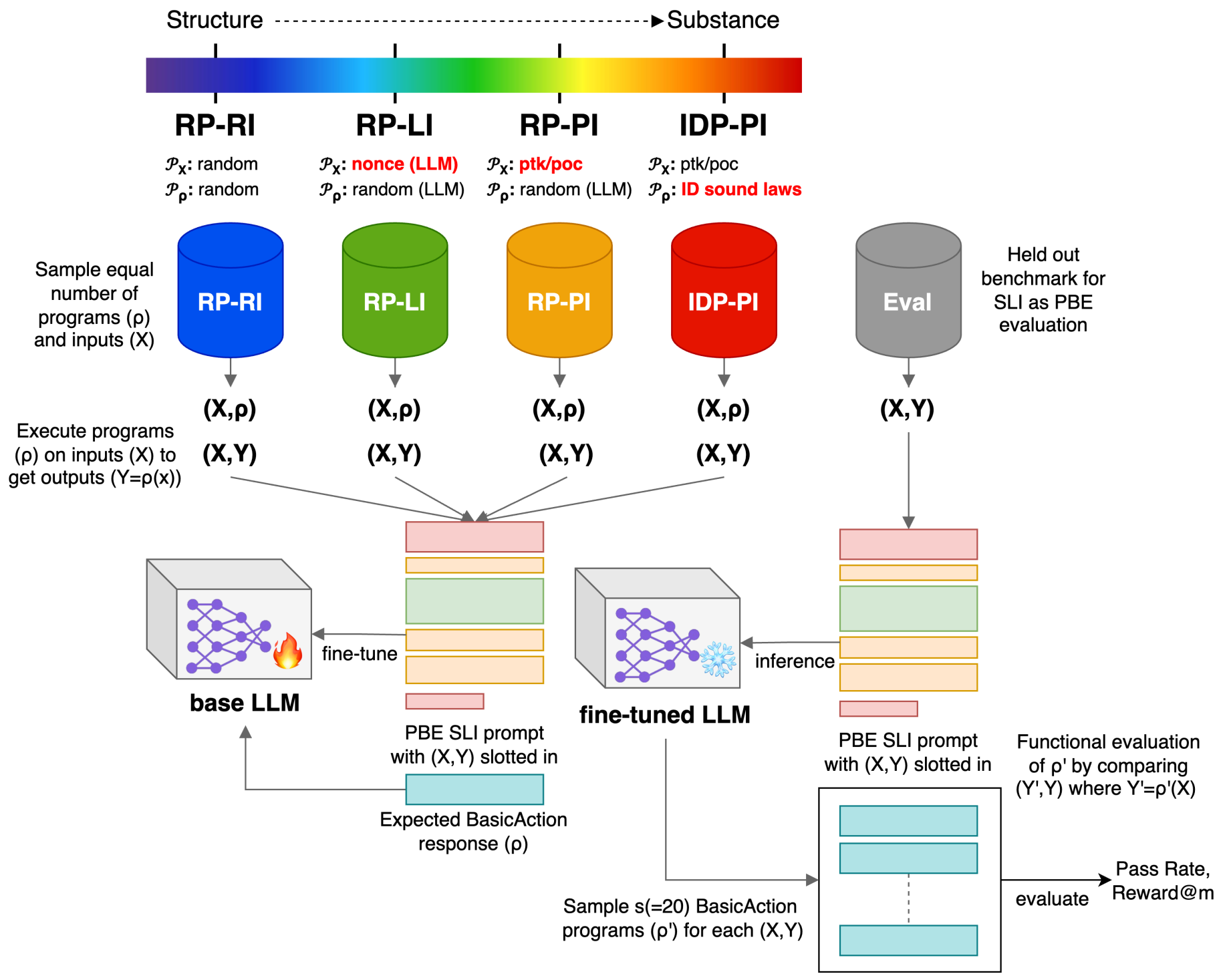

Figure 1: Prompt used for synthesizing single sound laws from the protoforms (inputs) and attested forms (outputs).

### 4.2 Synthetic Data Generation

We follow the general PBE data generation procedure discussed in section 2 while varying the input ($\mathcal{P}_X$) and program ($\mathcal{P}_\rho$) distributions, incrementally adding more substance. The implementation details for each synthetic data algorithm/condition are discussed below.

**RP-RI (Random Programs + Random Inputs):** In this setting, the sound laws are randomly sampled with contexts of one to three phones with a few (25%) having boundary-based conditions (word start, word end, not word start, or not word end), that perform one or more phone additions, deletions, or substitutions. For each sound law appropriate $N(= 50)$ inputs are randomly sampled such that at least $\frac{2}{5}N$ protoforms contain one or more occurrences of the context with $\frac{1}{10}N$ each beginning and ending with the context respectively (boundary locations), and $\frac{1}{10}N$ containing at least one and $\frac{1}{10}N$ containing at least two occurrences of the context in non-boundary locations.

**RP-LI (Random Programs + LLM Generated Nonce Inputs):** In this setting we prompt Codestral-22B with 5 seed example nonce words (WuggyCode, 2020) and programs that apply to them and prompt it to generate more inputs and programs. For each program and input pair, we add distractor inputs from other program-input pairs to ensure 50 examples, $(X, Y)$ per program, $\rho$. Programs that fail to affect inputs are filtered out.

**RP-PI (Random Programs + Protolanguage Inputs):** This setting is similar to the previous one, but instead of prompting the LLM to generate nonce inputs, it is directly prompted with 5 random words from Proto-Tangkhulic (ptk) or Proto-Oceanic (poc) (leading to two settings: RP-PI-ptk and RP-PI-poc) to generate sound laws that apply to them. Similar to the previous setting we add distractor inputs to the input pairs to ensure 50 examples per program and filter out programs that don't affect any inputs.

**IDP-PI (Index Diachronica Programs + Protolanguage Inputs):** In this setting we replace the LLM-generated programs with randomly sampled sound laws from the Index Diachronica (ID) (Index Diachronica Contributors, 2016) a database of sound laws from natural languages. We sample 50 random inputs from Proto-Tangkhulic or Proto-Oceanic (again leading to two settings: IDP-PI-ptk and IDP-PI-poc). Based on the sampled inputs, we find the longest common subsequences (LCS) between all pairs of inputs and sample a context from them with a probability proportional to the frequency of the LCS in the sampled inputs. This subsequence is treated as a context for the sound law program and is used to filter down to a set of applicable ID programs, out of which one is randomly selected.

### 4.3 Supervised Fine-tuning (SFT) on Synthetic Data

To compare the four synthetic data conditions (example inputs and programs shown in Table 3) on equal footing we train the open-source Magicoder-S-DS-6.7B (Wei et al., 2024) LLM on 2.5k instances $(X, Y, \rho)$, with 10% of the data being set aside for validation. We use the fine-tuning code published by the Magicoder creators[1] and train the model with a batch size of 1, max sequence length of 4500 and max length padding using an 80GB, A100 GPU for 3 epochs, evaluating the model every 200 steps, requiring almost three hours. We use the final checkpoint for evaluation. We selected Magicoder because of its relatively small size and publicly available fine-tuning script. Based on this study we determine the best condition for creating synthetic data and use it train a strong open source LLM for low resource SLI, fine-tuned from the base Magicoder model. We call this model **PySLICoder**.

## 5 Experiments

We use the SLI-as-PBE task to answer the question about the most useful training distribution for PBE raised in section 2. The end-to-end SLI task involves the induction of multiple sound laws (also called cascades) from paired inputs $X$ (protoforms) and paired outputs $Y$ (reflexes). However, this involves a search and ordering component which is not relevant to our study and can complicate the analysis. To remove the search component we create a "single law" version of the task where only a single sound law program ($\rho$) needs to be induced to map the inputs $X$ to the outputs $Y(=\rho(X))$. We split the experiments into two stages: 1) Comparing synthetic data conditions and 2) Benchmarking PySLICoder against other LLMs. For the first experiment, we fine-tune (details in section 4.3) Magicoder for each condition and find the best-performing condition. Then having determined the best condition, we train the Magicoder model on two variants of it, where we use Codestral-22B and GPT-4o, respectively, to generate the data and thus obtain two versions of PySLICoder. The subsequent sections describe the benchmark creation, models compared, and evaluation metrics for SLI.

### 5.1 Evaluation Benchmark Creation

Since we want to advance low-resource SLI, we use both real-world Polynesian and Tangkhulic sound change data. To construct a PBE dataset out of the sound change data, we started by manually deducing the most likely relative chronologies of sound change for the given paired proto-language headwords and descendant reflex forms in each language, assisted by the CFR-debugging system DiaSim (Marr and Mortensen, 2020). We converted the ordered rules of these into BasicAction functions to constitute the "gold" cascades with end-to-end accuracy of 75% or more for the development of Proto-Polynesian (Pol) into Hawaiian, Samoan, Niue, and Tongan, and of Proto-Tangkhulic (Ptk) into Huishu. To convert the multiple sound law cascades to the "single law" dataset (as discussed earlier in section 5), we iterate over the laws in the cascade, executing them and retaining the inputs and outputs at each step. The outputs are then fed as inputs to next sound law in the cascade. The input-output pairs for each sound law are constructed by keeping all the changed words from the intermediate inputs and outputs for that sound law and a small random amount of unchanged words as distractors. Following this process, we end up with a dataset of 17 Pol-Hawaiian, 5 Pol-Niue, 12 Pol-Samoan, 8 Pol-Tongan and 43 Ptk-Huishu sound law programs, with the min, max and median number of input-output examples being 11, 48 and 16, respectively.

### 5.2 Models

We compare the following open-source and closed source code LLMs with PySLICoder:

[1] https://github.com/ise-uiuc/magicoder/blob/main/src/magicoder/train.py

| Condition | Pass Rate | | | | | |
|---|---|---|---|---|---|---|
| | Pol-Hawaiian | Pol-Niue | Pol-Samoan | Pol-Tongan | Ptk-Huishu | Avg |
| RP-RI | 51.0 | 93.3 | 41.7 | 26.8 | 51.2 | 52.8 |
| RP-LI | 52.9 | 100 | 44.5 | 37.5 | 60.5 | **59.1** |
| RP-PI-ptk | 41.2 | 93.3 | 41.7 | 37.5 | 58.9 | 54.5 |
| RP-PI-poc | 54.9 | 80.6 | 61.1 | 37.5 | 58.1 | <u>58.4</u> |
| IDP-PI-ptk | 41.2 | 60 | 36.1 | 33.3 | 24.8 | 39.1 |
| IDP-PI-poc | 56.9 | 60 | 61.1 | 37.5 | 37.2 | 50.5 |

Table 1: Comparing pass rates achieved by Magicoder-S-DS-6.7B fine-tuned on equal amounts (2.5k) of synthetic data for each condition. Codestral-22B is used for all LLM-based synthetic data (RP-LI, RP-PI). All results are averaged across 3 runs. The "Avg" column denotes the arithmetic average of the individual pass rates for each language pair. The bolded and underlined results show the best and second-best numbers per column respectively.

**Open Source LLMs:** We evaluate recent, SOTA, small to medium sized (between 6.7B to 22B parameters) open-source instruction-tuned code LLMs trained with a variety of architectures including mixture-of-experts (MoE) models:

**Magicoder-6.7B-S-DS** (Wei et al., 2024) is instruction tuned using synthetic data with Evol-Instruct and OSS-Instruct with CodeLlama (CL) or DeepSeekCoder (DS) as the base model. We chose the DS version because of better, reported performance on coding benchmarks.

**DeepSeekCoder-7B-Instruct-v1.5** (Guo et al., 2024) is an open source LLM trained on high-quality project-level code with large context windows (16K tokens) and an infilling objective. We chose it because of its competitive performance on coding tasks despite being small.

**Qwen2.5-Coder-7B-Instruct** (Hui et al., 2024) is a recent, instruction-tuned LLM series of models that achieves SOTA performance across 10 diverse coding benchmarks. We chose it because of its recency and relative promise for coding tasks.

**DeepSeekCoder-v2-16B-Instruct** (Zhu et al., 2024) is a mixture-of-expert LLM trained for math & coding tasks in 338 programming languages, with a context length of 128K. We chose it because it is reported to be better than DeepSeekCoder-33B, which was used by (Li and Ellis, 2024).

**Codestral-22B-v0.1** (AI, 2024) is an open-source code LLM released by Mistral AI that supports 80+ programming languages. We chose it because of its coding abilities and larger parameter size.

**Closed Source LLM APIs:** We utilize OpenAI's GPT-4o a strong closed source LLM (Achiam et al., 2023; Hurst et al., 2024) to potentially get an upper bound on our PBE for SLI benchmark.

### 5.3 Evaluation Metrics

We evaluate the functional correctness of the LLM-generated sound laws by parsing out the BasicActions from the LLM response and executing them to obtain predicted reflexes and comparing them with the "gold" ones using the following metrics:

**Edit Distance Reward:** This metric is derived from (details in Appendix E.1) the reward function defined by (Luo, 2021). For each predicted program ($\rho$) the reward compares the predicted reflexes $s_{i,\rho}^{pred}$, ($i$ denoting the index of the instance in the dataset $\mathcal{D}$) against the ground truth reflexes $s_i^{target}$, where the initial ancestral forms are denoted by $s_i^{source}$ and ($dist$) is the aggregated Levenshtein edit distance between all input-output examples.

$$R(s_i^{source}, s_{i,\rho}^{pred}, s_i^{target}) = 1 - \frac{dist(s_{i,\rho}^{pred}, s_i^{target})}{dist(s_i^{source}, s_i^{target})}$$

The metric attains a maximum value of 1 when $s_{i,\rho}^{pred} = s_i^{target}$ or $dist(s_{i,\rho}^{pred}, s_i^{target}) = 0$. Also, the reward can take negative values, when $dist(s_{i,\rho}^{pred}, s_i^{target}) > dist(s_i^{source}, s_i^{target})$. We define reward@m as the average reward achieved by the top-$m$ sampled programs (denoted by $\rho \in \mathcal{P}^m$) (when sorted by the reward) and then aggregate it over all the instances $i \in \mathcal{D}$:

$$\text{reward@m} = \frac{1}{|\mathcal{D}|} \sum_{i \in \mathcal{D}} \sum_{\rho \in \mathcal{P}^m} R(s_i^{source}, s_{\rho,i}^{pred}, s_i^{target})$$

**Pass Rate** captures whether the model can produce a correct program, i.e., a program that is functionally equivalent to the "gold" program and achieves full reward within $s(=20)$ samples. In other words, percentage of instances $i$ with reward@1 =

1 or:

$$\text{pass_rate} = \frac{1}{|\mathcal{D}|} \sum_{i \in \mathcal{D}} \sum_{\rho \in \mathcal{P}^1} \mathcal{I}[R(s_i^{source}, s_{\rho,i}^{pred}, s_i^{target})]$$

Where $\mathcal{I}[x] = \begin{cases} 1, & \text{if } x = 1 \\ 0, & \text{otherwise} \end{cases}$

# 6 Results

## 6.1 Comparing Synthetic Data Conditions:

Table 1 shows the pass rate achieved by the four synthetic data conditions proposed in section 2. We run inference on each condition three times to reduce the effect of random variations, and sample $s(= 20)$ programs for each instance of the "single law" benchmark. RP-LI achieves the highest pass rate, followed closely by RP-PI-poc. The overall performance based on the pass rate is: (IDP-PI-ptk < IDP-PI-poc) < RP-RI < (RP-PI-ptk < RP-PI-poc) < RP-LI. We note that the choice of inputs among poc or ptk doesn't affect the trend of the synthetic data conditions: IDP-PI-ptk < RP-RI < RP-PI-ptk < RP-LI and IDP-PI-poc < RP-RI < RP-PI-poc < RP-LI or in other words IDP-PI < RP-RI < RP-PI < RP-LI. We also analyze the statistical significance of the differences between the conditions (for both the poc and ptk variants of PI), specifically looking at number of passing programs, reward per program, and pass rate. We run a paired Wilcoxon signed rank test (Wilcoxon, 1992) across the three runs of the compared conditions on the 85 data points and 20 samples ($85 \times 3 \times 20 = 5100$ values) for the reward per program and the passing programs comparison (i.e. programs with a reward of 1) and $85 \times 3 = 255$ values for pass rate (i.e. any of the 20 sampled programs has a reward of 1). Table 5 and compares all the fine-tuning conditions with the base Magicoder model. Table 6 and 7 show the statistical significance of the input and program distribution manipulations we do in the conditions for the poc and ptk cases.

## 6.2 Comparing PySLICoder with LLMs:

The best synthetic data condition (RP-LI) is used to train PySLICoder. We propose two variants of PySLICoder: PySLICoder-RP-LI-codestral which is trained on RP-LI data generated with Codestral-22B and PySLICoder-RP-LI-gpt-4o which is trained on RP-LI data generated by GPT-4o. We compare both of these variants with other coding LLMs in Table 2. PySLICoder-RP-LI-gpt-4o attains the best overall performance among open-source models but is worse than GPT-4o. PySLICoder-RP-LI-codestral is slightly worse than Codestral-22B but achieves the best pass rate on the Ptk-Huishu language pair among open-source models.

# 7 Discussion

The results from Table 2 suggest that having an input distribution somewhat similar to the evaluation data (like poc words (PI) and nonce words (LI)) helps performance (RP-PI > RP-RI and RP-LI > RP-RI) but adding additional biases to the sound law programs can be detrimental (IDP-PI < RP-PI and IDP-PI < RP-RI).

**Effect of Input Distribution:** We analyze the impact of the input distribution trend (RI < PI < LI) by looking at the statistical significance of RP-RI < RP-PI-poc < RP-LI (Table 6) and RP-RI < RP-PI-ptk < RP-LI (Table 7). For poc we notice a statistically significant increase in reward per program in all cases, and passing programs for PI < LI, but there is not statistically significant difference among the pass rates. For ptk we notice a statistically significant trend for both reward per program and passing programs but not for pass rates. The trends suggest that having more realistic inputs helps, but only up to a point. Inputs that are not exactly words from the vocabulary, but are still 'pronounceable', lead to the best rewards per program and passing programs.

**Effect of Program Distribution:** We analyze the impact of inductive biases in the programs by comparing IDP-PI-poc < RP-PI-poc (Table 6) and IDP-PI-ptk < RP-PI-ptk (Table 7). Here for both poc and ptk we notice a significant difference in reward per program. For pass rate and passing programs, we notice a significant difference for ptk but not for poc. Thus IDP can negatively impact the performance, including the pass rate. We also compare IDP-PI-poc < RP-RI and IDP-PI-ptk < RP-RI to check if IDP-PI is worse than the lowest bias condition RP-RI. We notice conflicting results between ptk and poc. For ptk there is a significant difference for all three properties but no difference for poc. So IDP-PI is worse than RP-RI for ptk but not poc. This might stem from a lack of program diversity in IDP-PI-ptk (discussed in Appendix D).

**Impact of Language Vocabulary for PI (ptk vs poc):** We notice a consistent trend of conditions with poc outperforming their ptk counterparts (IDP-PI-ptk < IDP-PI-poc and RP-PI-ptk < RP-PI-poc). We analyze the statistical significance of these trends and notice a significant difference in all three properties for IDP-PI but not RP-RI as shown in Table 8. We believe this is again due to a lack of program diversity in IDP-PI-ptk.

| Model | Pass Rate | | | | | |
|---|---|---|---|---|---|---|
| | Pol-Hawaiian | Pol-Niue | Pol-Samoan | Pol-Tongan | Ptk-Huishu | Avg |
| Magicoder-S-DS-6.7B | 2.0 | 0 | 5.6 | 0 | 4.3 | 2.4 |
| DeepSeekCoder-7B-Instruct-v1.5* | 11.8 | 20 | 16.7 | 12.5 | 7 | 13.6 |
| Qwen2.5-Coder-7B-Instruct* | 23.5 | 40 | 8.3 | 12.5 | 21 | 21.1 |
| DeepSeekCoder-v2-16B-Instruct* | 0 | 40 | 16.7 | 0 | 9.3 | 13.2 |
| Codestral-22B | 54.9 | **100** | 55.6 | 50 | 50.4 | 62.2 |
| PySLICoder-RP-LI-codestral | 52.9 | **100** | 44.5 | 37.5 | 60.5 | 59.1 |
| PySLICoder-RP-LI-gpt-4o | 82.35 | **100** | 68.4 | 37.5 | 58.1 | 68.4 |
| GPT-4o | **90.2** | **100** | **72.2** | **75** | **72.9** | **82.1** |

Table 2: Pass Rate of LLMs on SLI as PBE. All results except the starred models are averages across 3 runs. * - results from a single run. The "Avg" column denotes the simple arithmetic average of the individual pass rates for each language pair.

**Effect of Fine-tuning**: Table 5 shows a statistically significant increase in the pass rate, reward per program, and passing programs for all the fine-tuning conditions compared to the base Magicoder showing the effectiveness of the fine-tuning.

**PySLICoder vs other Coding LLMs:** The results in Table 2 show that PySLICoder-RP-LI-gpt-4o achieves a 6% better pass rate than Codestral-22B ($p<0.01$), higher reward per program ($p<0.0001$) and passing programs ($p<0.0001$), making it the strongest open source LLM for SLI while having a third of the parameters of Codestral. The results also confirm the observations of (Li and Ellis, 2024) about fine-tuning being necessary for PBE as most open-source LLMs except Codestral-22B achieve poor pass rates. Finally, we note that both PySLICoder variants perform worse than the models used to generate their training data. We believe this might be due to the 2.5k fine-tuning samples being insufficient to mimic their performance.

## 8 Conclusions and Future Work

In this work, we presented a novel programming by examples (PBE) formulation of the task of sound law induction (SLI) for diachronic linguistics. We created a PBE evaluation benchmark for SLI in low-resource settings to evaluate state-of-the-art LLMs of code. Then inspired by recent work on PBE with LLMs (Li and Ellis, 2024; Fu et al., 2024) as well as the age-old **structure and substance debate** (Boye and Engberg-Pedersen, 2015) in linguistics we conceptualized a framework for generating synthetic data with controllable amounts of inductive bias in the input and program distributions. Using our framework, we proposed four approaches spanning the spectrum of inductive biases and conducted a controlled experiment to answer whether "structure" alone is enough for SLI or one can benefit from adding more "substance". Our results showed that a **middle of the road approach is best** with more substantive inputs being helpful, but greater biases in programs at the cost of diversity being harmful. Having determined that we used the best condition RP-LI to generate training data with Codestral-22B and GPT-4o and trained two "PySLICoder" models which achieved the best performance among open source models with the least number of parameters.

However, there are some limitations of this study like the small size of the benchmark due to tackling low resource scenarios. In future work, we plan to expand the evaluation benchmark with more language pairs for a comprehensive evaluation of SLI. We also plan to explore scaling up the synthetic data to analyze performance improvement and saturation, if the ordering of the conditions changes, and if the fine-tuned model can surpass the model used to generate the training data. Additionally, we also plan to explore if the findings in this study hold when we vary the models to be fine-tuned from Magicoder to Qwen2.5-Coder-7B, DeepSeekCoder–7B, etc., and if we can use them to develop a stronger version of PySLICoder. Finally, we also plan to explore if our findings generalize to other PBE domains like general string manipulation and data wrangling (Gulwani, 2016).

## Limitations

- **Small Evaluation Benchmark:** Since our aim is to be sample efficient and support low-resource settings we end up with a small benchmark of 85 instances. However to deal with the small size we ran the inference with each synthetic data condition, important baselines (Codestral-22B, Magicoder and GPT-4o) as well as PySLICoder three times to reduce variance in the results and report the averaged results. We also use all runs to determine statistical significance making the effective size of the dataset $85 \times 3 = 255$. Also, we sample $s = 20$ programs per instance which leads to $255 \times 2 = 5100$ values for comparing passing programs and reward per program allowing us to observe statistically significant differences across most of the synthetic data conditions. We believe the effective dataset size of 255 is reasonable as several popular code generation benchmarks like HumanEval (Chen et al., 2021) also only have a few hundred samples.

- **Potential Bias in Evaluation Benchmark:** A potential issue in our evaluation benchmark is that we have four language pairs with Proto Polynesian (pol) as the protolanguage which could favor the settings with the poc vocab over the ptk vocab in the downstream SLI performance. We tried to mitigate this with a weighted average version of the pass rate. However, in such a scenario the performance ended up being heavily biased towards models that perform well on just the Ptk-Huishu language pair because it has 43/85 instances. Thus we chose the simple average over the weighted average because we wanted to give a more equitable importance to each language pair rather than just favoring Ptk-Huishu. Also we observed that the weighted average still didn't affect the poc vs ptk trend because the models trained on the ptk vocab (IDP-PI-ptk, RP-PI-ptk) instead of poc (IDP-PI-poc, RP-PI-poc) don't always do better on Ptk-Huishu compared to the poc counterpart.

- **Limited Computational Budget for Program Sampling:** While (Li and Ellis, 2024) kept sampling from the LLM till they were able to achieve a passing program, we adopted a more budget-constrained approach to PBE where only sampled s(=20) programs for each dataset instance. The reason for this constrained sample budget was to reduce the inference time (which can take several minutes already) and represent a more realistic and constrained setting where the end-user has limited time/patience. While this could be arguably considered a limitation of our work we argue that our evaluation setup is more realistic, time-efficient, and promotes the development of more sample PBE approaches.

- **Computational Demands:** The use of Large Language Models (LLMs) requires considerable computational resources. Both fine-tuning and running LLMs demand extensive memory and GPU capabilities.

- **Fine-tuning more models:** While we recognize that we did all the fine-tuning experiments on a single model Magicoder-6.7B which we picked due to its small size, public fine-tuning code and ease of fine-tuning. One natural question here could be if the observations related to the fine-tuning data would hold for other similarly smaller code models as well. We plan to explore them in future work to see if the trend holds up for other models as well (e.g. DeepSeekCoder-v1.5-7B or Qwen2.5-Coder-7B).

- **Context Length Constraints:** LLMs are limited by the amount of information they can process at once. Our PySLICoder model, which is based on the Magicoder model had a much smaller context length compared to the latest models like DeepSeekCoder-V2 which have context lengths of 128K. Prompts with a large number of input-output examples combined with our verbose prompting format could exceed the context length of Magicoder/PySLICoder. In future work, we plan to develop long context versions of PySLICoder by using DeepSeekCoder-v2-Instruct-Lite MoE model with 128K context length as the base model and fine-tuning it on the synthetic data.

- **Choice of Magicoder for fine-tuning experiments** Based on the limitations acknowledged above we just wanted to re-iterate that while there were better models to be chosen instead of Magicoder, we wanted to ac-knowledge that many of these models came out after we formulated the fine-tuning experiments and based on aspects like parameter size, ease of fine-tuning and performance on coding benchmarks Magicoder-6.7B was one of the best choices.

## Ethics Consideration

In developing Large Language Models (LLMs) for sound law induction, it is imperative to ensure that the synthetic data utilized for fine-tuning is devoid of biases that could distort the generated sound laws. This necessitates a careful consideration of linguistic diversity and the avoidance of overrepresentation of specific language families or phonetic patterns. Moreover, using pre-existing linguistic data or frameworks in creating synthetic data must be approached with respect for intellectual property rights, ensuring proper attribution and consent from original researchers. Lastly, the automation of sound law induction presents a transformative potential for historical linguistics, promising to alleviate cognitive and temporal demands. However, balancing the benefits of automation and the invaluable role of human expertise and interpretative analysis in the field is crucial, safeguarding the nuanced understanding of linguistic evolution.

## Generative AI Assistant Use

We **didn't use** any generative AI assistants except for verifying punctuation while writing this paper.

# A Conceptual Framework

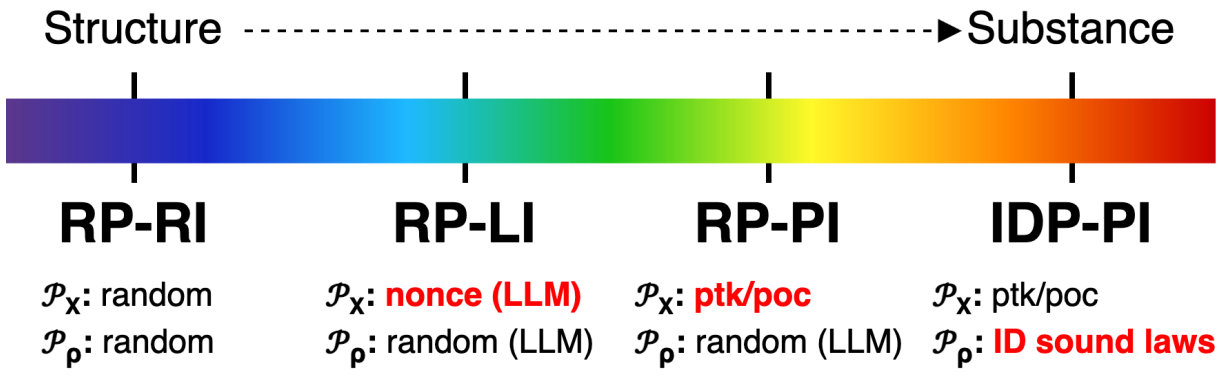

Figure 2: The four fine-tuning conditions introduced in this paper depicted along the spectrum of structure vs substance.

# B Method Details

## B.1 BasicAction Class Definition

The BasicAction class is used to represent the phonological transformations behind the sound laws. These transformations move from left to right over the list of tokenized phones (where the start and end are marked by a special token '#' and separated by a special token '@') in a word to match an environment specified as a list of predicates, where each predicate is a boolean function that matches to a set of phones (e.g. the is_nothing function matches to the '@' separator token). When a match is found deletions, substitutions (replacing phones with other phones based on a mapping), or insertions (mapping/replacing a phone with multiple phones) are performed at the set of relative positions specified in the list of change_pos according to the corresponding mapping function passed in the list mapping_fn.

An example BasicAction object showing the format of the predicates, change positions, and mappings is shown in Figure 4. An important implementation detail of the BasicAction is suppression of self-feeding (when rules can create the environment for their own application) which is achieved by a two-stage process of first detecting all locations/sites where an environment match is found and the modifying the appropriate phones in the second stage.

## B.2 Data Generation Prompts

We utilize the following prompts for generating data with LLMs for the RP-LI and RP-PI conditions:

**RP-LI:** We use the prompt shown below which first describes the BasicAction used and then gives instructions on generating the nonce word inputs as well as some examples of random programs and nonce words that apply to those programs to teach the LLM how to generate random programs and determine what nonce inputs would be affected by it.

You are going to be working with BasicAction a class used to represent sound laws that transform a list of source words into target words. A Basic Action is an object of the class BasicAction where the list `predicates` describes a context window where the change happens the integer `change_pos` describes the position of the character to be changed, and the function `mapping_fn` describes the substitution, deletion or addition to be done.

A Basic Action is applied to each source word which is a sequence of phonemes. Note that the phoneme sequence is first preprocessed by adding a '#' at the beginning and end of the sequence, and adding a '@' between each phoneme. Therefore, you need to consider the '#' and '@' when you write the predicates and mapping function. Namely remember to separate any consecutive predicates related to regular phonemes with `lambda x: x == '@'` placeholder matching predicates.

Here are some examples of how actions can be implemented, and some nonsense english like word inputs where each of them applies. Extend the list given here with your own diverse example actions and inputs where they apply.

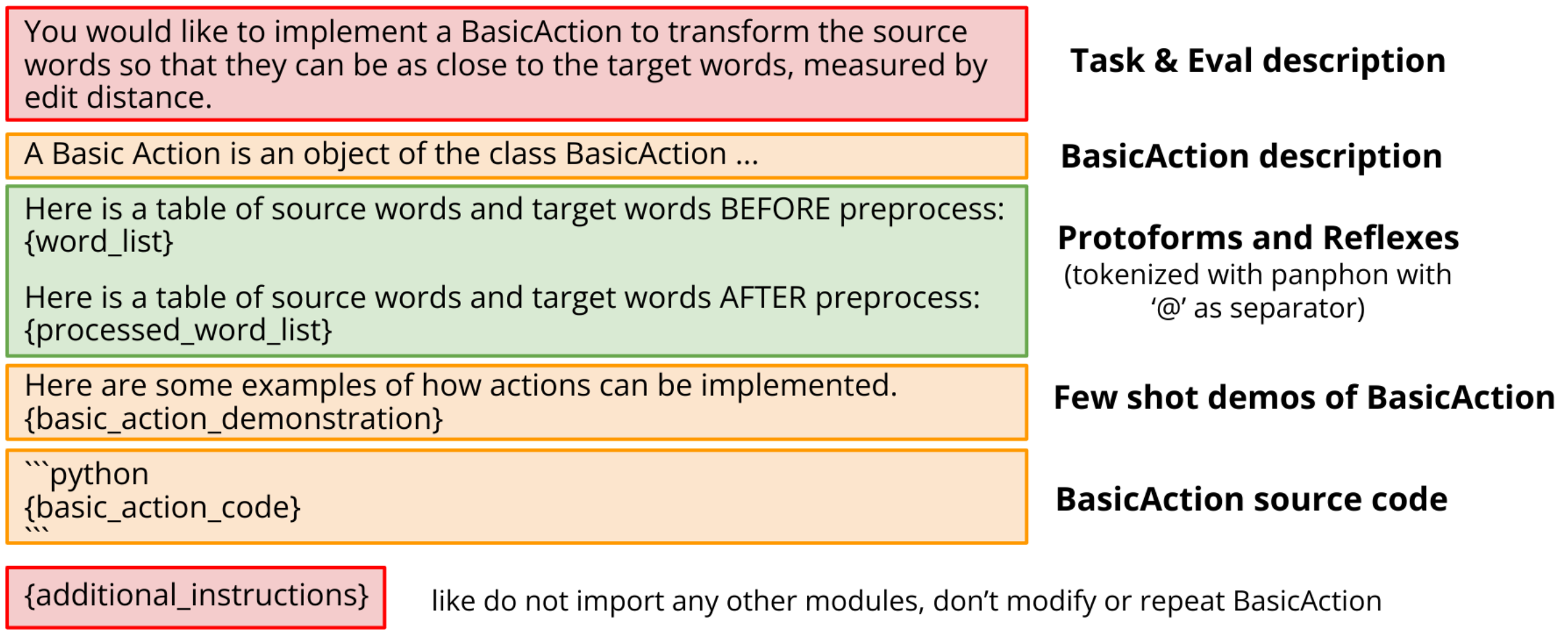

Figure 3: Prompt used for synthesizing single sound laws from the protoforms (inputs) and attested forms (outputs).

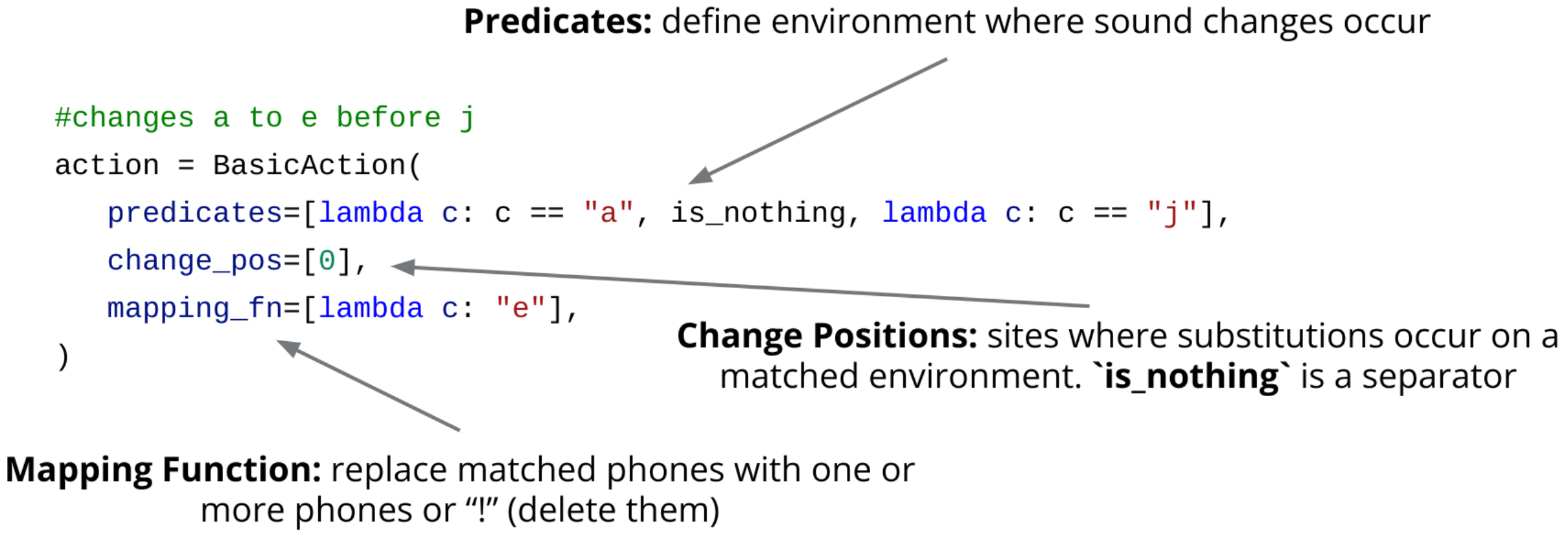

Figure 4: Instantiation of the BasicAction class to represent a sound law. This example shows a rule where "a" goes to "e" when it occurs before a "j". The predicates match to an environment of "a@j" where '@' is the separator, and then the first character of the environment or "a" goes to "e" as described by the change_pos and mapping_fn. In other words it represents the rule $a > e \backslash _j$

An action that changes a word-initial 'a' to 'e' can be written as follows:
```python
action = BasicAction(predicates=[lambda x: x == '#', lambda x: x == '@', lambda x: x == 'a'], change_pos=[2], mapping_fn=[lambda x: 'e'])
nonce_inputs = ['apolated', 'acklay', 'anspeckonre']
```
where the first predicate checks if it's the beginning of a word, the second predicate is a placeholder that indicates the gap between two phonemes, and the third predicate checks if the first phoneme is 'a'.

An action that deletes 's' in a context window of 'sal', 'sat' or 'sav' can be written as follows:
```python
action = BasicAction(predicates=[lambda x: x == 's', lambda x: x == '@', lambda x: x == 'a', lambda x: x == '@', lambda x: x in ['l', 't', 'v']], change_pos=[0], mapping_fn=[lambda x: '!'])
nonce_inputs = ['neyingersalved', 'savolcomish', 'sataphier']
```

where the first predicate checks if the first phoneme is 's', the second predicate is a placeholder that indicates the gap between two phonemes, the third predicate checks if the third phoneme is 'a', the fourth is also a placeholder like the second predicate and the last predicate checks if the phoneme is an `l`, `t` or `v`.

An action that adds 'b' after an 'a' in a context window of 'a' not at the end can be written as follows:
```python
action = BasicAction(predicates=[lambda x: x == 'a', lambda x: x == '@', lambda x: x != '#'], change_pos[0], mapping_fn=[lambda x: x+'b'])
nonce_inputs = ['nerocеash', 'avockdentivery', 'tranalk', 'albalmenests']
```
where the first predicate checks if the first phoneme is 'a', the second predicate is a placeholder that indicates the gap between two phonemes, and the third predicate checks if the third phoneme is not a boundary character or '#'.

| Condition | Inputs | Programs | Outputs | Explanation |
|---|---|---|---|---|
| RP-RI | mztkeɨsthwsp$^{w}$j<br>ðꝺtrnegv<br>wotzhtŋ | k>∅ / _#<br>w>∅ / _#<br>ŋ>∅ / _# | mzteɨsthsp$^{w}$j<br>ðꝺtrnegv<br>otzht | Back consonant (k,w,ŋ) gets deleted |
| RP-LI | sunt<br>einstein<br>tapere | t>d / _# | sund<br>einstein<br>tapere | ’t’ goes to ’d’ before a boundary |
| RP-PI-ptk | t$^{h}$um<br>sam | m>n / _# | t$^{h}$un<br>san | ’m’ goes to ’n’ before a boundary |
| IDP-PI-poc | talun<br>tumpul<br>suat | u>o / _C | talon<br>tompol<br>suat | ’u’ goes to ’o’ before consonant (C) |

Table 3: Comparison of the four types of synthetic data generated. Each condition gains more substance as we go from top (RP-RI) to bottom (IDP-PI-poc). RP-RI has random, unpronounceable inputs and randomly sampled string-manipulation programs, RP-LI uses Codestral-22B to generate pseudoword inputs and programs. RP-PI-ptk uses actual Proto-Tangkhulic words and prompts Codestral-22B with them to generate programs, while IDP-PI-poc uses real Proto-Oceanic words along with sound laws sampled from Index Diachronica.

````
An action that adds 'b' before an 'a' in a context
window of 'a' not at the end can be written as
follows:
```python
action = BasicAction(predicates=[lambda x: x ==
'a', lambda x: x == '@', lambda x: x != '#'],
change_pos[0], mapping_fn=[lambda x: 'b'+x])
nonce_inputs = ['neroceash', 'avockdentivery',
'tranalk', 'albalmenests']
```
where the first predicate checks if the first phoneme
is 'a', the second predicate is a placeholder that
indicates the gap between two phonemes, and the
third predicate checks if the third phoneme is not a
boundary character or '#'.

An action that changes a vowel at the end to 'p'
can be written as follows:
```python
action = BasicAction(predicates=[lambda x: x in
['a','e','i','o','u'], lambda x: x == '@', lambda x: x
== '#'], change_pos[0], mapping_fn=[lambda x:
'p'])
nonce_inputs = ['sureau', 'niatle', 'possma',
'wanlini', 'gargro']
```
where the first predicate checks if the first phoneme
is a vowel, the second predicate is a placeholder
that indicates the gap between two phonemes, and
the third predicate checks if the first phoneme is not
a boundary character or '#'.

Now write more actions that follow this format:
````

**RP-PI:** For this setting we use a prompt that gives the LLM a set of sampled inputs first and then asks it produce a BasicAction that applies to it based on few-shot examples of how to do so.

````
You are going to be working with BasicAction a
class used to represent sound laws that transform a
list of source words into target words. A Basic
Action is an object of the class BasicAction where
the list `predicates` describes a context window
where the change happens the integer `change_pos`
describes the position of the character to be
changed, and the function `mapping_fn` describes
the substitution, deletion or addition to be done.

A Basic Action is applied to each source word
which is a sequence of phonemes. Note that the
phoneme sequence is first preprocessed by adding a
'#' at the beginning and end of the sequence, and
adding a '@' between each phoneme. Therefore, you
need to consider the '#' and '@' when you write the
predicates and mapping function. Namely remember
to separate any consective predicates related to
regular phonemes with `lambda x: x == '@'`
placeholder matching predicates.

Here are some examples of how actions can be
implemented to transform every word in a given set
of nonsense words. Your task will be to propose
multiple actions for the given set of nonsense words.

Example nonsense words: ['san', ' an', 'lam', 'wam',
' ap', ' a', 't a', 'lar', 'tsar', 'ha']

An action that changes an 'a' to 'e' can be written
as follows:
```python
action = BasicAction(predicates=[lambda x: x ==
'a'], change_pos=[0], mapping_fn=[lambda x: 'e'])
```

An action that adds an 'e' before 'a' can be written
as follows:
```python
````
````

~~~
action = BasicAction(predicates=[lambda x: x ==
'a'], change_pos=[0], mapping_fn=[lambda x:
'e'+x])
```

An action that adds an 'e' after 'a' can be written
as follows:
```python
action = BasicAction(predicates=[lambda x: x ==
'a'], change_pos=[0], mapping_fn=[lambda x:
x+'e'])
```

Example nonsense words: ['pam', ' am', 'vam',
'ham', 'sam', 't am', 'cam', 'tsam', 'k am', 'ram']

An action that deletes an 'm' after 'a' can be
written as:
```python
action = BasicAction(predicates=[lambda x: x ==
'a', lambda x: x == '@', lambda x: x == 'm'],
change_pos=[2], mapping_fn=[lambda x: '!'])
```

An action that deletes a word final 'm' after 'a' can
be written as:
```python
action = BasicAction(predicates=[lambda x: x ==
'a', lambda x: x == '@', lambda x: x == 'm',
lambda x: x == '@', lambda x: x == '#'],
change_pos=[2], mapping_fn=[lambda x: '!'])
```

Example nonsense words: ['tap', 'cap', 'cep', 'tep',
'tip', 'tup', 'cup']

An action that deletes a vowel before a word-final
'p' can be written as follows:
```python
action = BasicAction(predicates=[lambda x: x in
['a','e','i','o','u'], lambda x: x == '@', lambda x: x
== 'p', lambda x: x == '@', lambda x: x == '#'],
change_pos=[0], mapping_fn=[lambda x: '!'])
```

An action that replaces a vowel with ' ' when
flanked by 't' or 'c' in front and 'p' in the back can
be written as follows:
```python
action = BasicAction(predicates=[lambda x: x in
['t','c'], lambda x: x == '@', lambda x: x in
['a','e','i','o','u'], lambda x: x == '@', lambda x: x
== 'p'], change_pos=[2], mapping_fn=[lambda x:
' '])
```

Example nonsense words: {input_words}
~~~

Although the prompts are very large we report them verbatim to aid reproducibility.

# C More Results

Table 4 shows the reward@1 and reward@3 attained by the evaluated LLMs, each Magicoder variant (trained on each synthetic data condition) from the fine-tuning study as well as the two proposed PySLICoder models. Please note that PySLICoder-RP-LI-Codestral is the Magicoder (RP-LI) condition.

# D Qualitative Comparison of Synthetic Data

## D.1 RP-PI-ptk V/s RP-PI-poc

Referencing Table 9, which includes examples that fairly represent the different types of rules observed in each setting. These examples provide valuable insights into the characteristics of the rules within each context.

Majority of the rules in PTK focus on identifying a single character in a word, without being selective about its position, and then replacing it with another character or occasionally adding a character after the matched one (rows 1,2: column 1). There are a few rules, however, which check for the position of the character in the word (row 3: column 1) and also some other rules which check for multiple characters i.e. a set of characters simultaneously in one rule (row 4: column 1).

In the POC setting, we observed rules that check for the presence of specific character sequences, i.e., multiple characters occurring in a prescribed order (rows 2, 3: column 2). Additionally, there are severa; rules that check for the absence of a set of characters, such as vowels (rows 2, 4: column 2). However, checking for long character sequences and the absence of character sets is something we did not observe in the PTK setting.

## D.2 IDP-PI-poc V/s IDP-PI-ptk

In this section, we will refer to Table 10, which provides examples illustrating the various types of rules encountered in each setting.

In the PTK setting, by examining column 1 of the table, we observe that the rules check for the presence of a single character in the word. Subsequently, the rules suggest replacing this character with a sequence of one, two, or, in some cases, three characters.

Observing the rules corresponding to the POC setting (column 2), we notice simple rules that check for the presence of a character anywhere in the word and suggest replacing it with another character (rows 1,2: column 2). Additionally, there are more complex rules that check for character sequences and, within those sequences, use functional definitions to identify the presence of vowels or consonants.

| Model | Atr-Hawaiian | | Atr-Niue | | Atr-Samoan | | Atr-Tongan | | Ptk-Huishu | |
|---|---|---|---|---|---|---|---|---|---|---|
| | R@1 | R@3 | R@1 | R@3 | R@1 | R@3 | R@1 | R@3 | R@1 | R@3 |
| Magicoder-S-DS-6.7B | 0.0712 | 0.0321 | 0.1333 | 0.0508 | 0.1095 | 0.0374 | 0.0301 | 0.01 | 0.0989 | 0.0433 |
| DeepSeekCoder-7B-Instruct-v1.5* | 0.2258 | 0.1797 | 0.2714 | 0.2 | 0.2045 | 0.1222 | 0.125 | 0.0833 | 0.1568 | 0.0986 |
| Qwen2.5-Coder-7B-Instruct* | 0.3459 | 0.2229 | 0.5714 | 0.3238 | 0.1726 | 0.1204 | 0.1875 | 0.0625 | 0.5024 | 0.3389 |
| DeepSeekCoder-v2-16B-Instruct* | 0.1177 | 0.0711 | 0.5 | 0.3667 | 0.2444 | 0.1472 | 0.0625 | 0.0625 | 0.1286 | 0.1128 |
| Codestral-22B | **0.7355** | **0.6504** | **1** | **0.9698** | **0.6917** | **0.5624** | **0.6706** | **0.487** | **0.7651** | **0.6400** |
| Magicoder (RP-RI) | 0.7072 | 0.6219 | 0.9333 | 0.8915 | 0.6164 | 0.5838 | 0.4780 | 0.4092 | 0.7525 | 0.724 |
| PySLICoder-RP-LI-codestral | 0.7736 | **0.7723** | **1** | **1** | 0.6758 | 0.6619 | 0.6051 | 0.5751 | **0.8527** | **0.8165** |
| Magicoder (RP-PI-ptk) | 0.5349 | 0.4594 | 0.9333 | 0.9000 | 0.4841 | 0.4815 | 0.5327 | 0.5149 | 0.8083 | 0.7929 |
| Magicoder (RP-PI-poc) | **0.8022** | 0.7416 | 0.873 | 0.8704 | **0.7255** | **0.7222** | **0.6696** | **0.6333** | 0.7889 | 0.7421 |
| Magicoder (IDP-PI-ptk) | 0.5299 | 0.4874 | 0.6 | 0.5778 | 0.4583 | 0.3769 | 0.4117 | 0.3555 | 0.6126 | 0.5758 |
| Magicoder (IDP-PI-poc) | 0.6280 | 0.5348 | 0.7846 | 0.7846 | 0.6806 | 0.5421 | 0.5714 | 0.4405 | 0.6123 | 0.5925 |
| PySLICoder-RP-LI-gpt-4o | 0.9481 | 0.9472 | 1 | 1 | 0.8261 | 0.7933 | 0.7371 | 0.7236 | 0.7796 | 0.7669 |
| GPT-4o | **0.9456** | **0.8813** | **1** | **1** | **0.7936** | **0.7237** | **0.8274** | **0.797** | **0.8789** | **0.8563** |

Table 4: The Rewards@1 (R@1) and Rewards@3 (R@3) attained by various LLMs on sound law induction. All results except the starred models are averages across 3 runs. * - results from a single run. The bolded results show the best performance in every section (open source, fine-tuned and closed-source models).

| Comparison | Reward per Program | | Passing Programs | | Pass Rate | |
|---|---|---|---|---|---|---|
| | p-value | statistic | p-value | statistic | p-value | statistic |
| no-finetune <RP-RI | p<0.0001 | 1.40e+06 | p<0.0001 | 1.89e+03 | p<0.0001 | 192 |
| no-finetune <RP-LI | p<0.0001 | 2.59e+05 | p<0.0001 | 4.60e+03 | p<0.0001 | 296 |
| no-finetune <RP-PI-ptk | p<0.0001 | 9.79e+05 | p<0.0001 | 2.39e+03 | p<0.0001 | 204 |
| no-finetune <RP-PI-poc | p<0.0001 | 2.65e+05 | p<0.0001 | 1.92e+03 | p<0.0001 | 214 |
| no-finetune <IDP-PI-ptk | p<0.0001 | 1.57e+06 | p<0.0001 | 2.98e+03 | p<0.0001 | 270 |
| no-finetune <IDP-PI-poc | p<0.0001 | 1.09e+06 | p<0.0001 | 0 | p<0.0001 | 0 |
| no-finetune <RP-LI-gpt-4o | p<0.0001 | 1.11e+05 | p<0.0001 | 2.44e+03 | p<0.0001 | 163 |

Table 5: Does fine-tuning produce a significant improvement? (PySLICoder vs base Magicoder comparison). Wilcoxon Signed Rank test results show that each fine-tuning setting achieves a statistically significant improvement over the base Magicoder runs for reward per program, number of passing programs and pass rate. Note we use significance level $\alpha = \frac{0.05}{m} = \frac{0.05}{7} = 0.007$ in accordance with the Bonferroni correction (Weisstein, 2004) to account for the 7 comparisons.

| Comparison | Reward per Program | | Passing Programs | | Pass Rate | |
|---|---|---|---|---|---|---|
| | p-value | statistic | p-value | statistic | p-value | statistic |
| RP-RI <RP-PI-poc | **p<0.0001** | 2.95e+06 | p>0.05 | 6.05e+05 | p>0.05 | 2.48e+03 |
| RP-PI-poc <RP-LI | **p<0.0001** | 1.90e+06 | **p<0.0001** | 6.10e+05 | p>0.05 | 3.36e+03 |
| RP-RI <RP-LI | **p<0.0001** | 2.27e+06 | **p<0.0001** | 6.64e+05 | p>0.05 | 3.43e+03 |
| IDP-PI-poc <RP-PI-poc | **p<0.01** | 3.36e+06 | p>0.05 | 9.00e+05 | p>0.01 | 3.99e+03 |
| IDP-PI-poc <RP-RI | p>0.05 | 4.26e+06 | p>0.05 | 8.49e+05 | p>0.05 | 3.10e+03 |

Table 6: Statistical significance Wilcoxon Signed Rank tests performed with the poc variants of IDP-PI and RP-PI (significance level $\alpha = \frac{0.05}{5} = 0.01$ after Bonferroni correction Weisstein (2004)). We highlight statistically significant observations.

The rules for the POC setting exhibit greater diversity in the operations they can model compared to the PTK setting. This increased diversity may contribute to the superior performance of POC over PTK in the IDP setting.

Another key observation is that, compared to the RP setting, the rules in the IDP setting tend to be less diverse.

| Comparison | Reward per Program | | Passing Programs | | Pass Rate | |
|---|---|---|---|---|---|---|
| | p-value | statistic | p-value | statistic | p-value | statistic |
| RP-RI <RP-PI-ptk | **p<0.0001** | 3.32e+06 | **p<0.0001** | 6.61e+05 | p>0.05 | 2.83e+03 |
| RP-PI-ptk <RP-LI | **p<0.0001** | 3.01e+06 | **p<0.0001** | 1.08e+06 | p>0.05 | 3.81e+03 |
| RP-RI <RP-LI | **p<0.0001** | 2.27e+06 | **p<0.0001** | 6.64e+05 | p>0.05 | 3.43e+03 |
| IDP-PI-ptk <RP-PI-ptk | **p<0.0001** | 2.74e+06 | **p<0.0001** | 5.07e+05 | **p<0.0001** | 2.66e+03 |
| IDP-PI-ptk <RP-RI | **p<0.0001** | 3.32e+06 | **p<0.0001** | 4.94e+05 | **p<0.0001** | 2.50e+03 |

Table 7: Statistical significance Wilcoxon Signed Rank tests performed with the ptk variants of IDP-PI and RP-PI (significance level $\alpha = \frac{0.05}{5} = 0.01$ after Bonferroni correction Weisstein (2004)). We highlight statistically significant observations.

| Comparison | Reward per Program | | Passing Programs | | Pass Rate | |
|---|---|---|---|---|---|---|
| | p-value | statistic | p-value | statistic | p-value | statistic |
| RP-PI-ptk <RP-PI-poc | **p<0.0001** | 3.03e+06 | **p<0.0001** | 5.27e+05 | **p<0.001** | 2.40e+03 |
| ID-PI-ptk <IDP-PI-poc | p>0.05 | 3.59e+06 | p>0.05 | 9.02e+05 | p>0.05 | 3.24e+03 |

Table 8: Statistical significance Wilcoxon Signed Rank tests between ptk and poc variants of IDP-PI and RP-PI (significance level $\alpha = \frac{0.05}{2} = 0.025$ after Bonferroni correction Weisstein (2004)). We highlight statistically significant observations.

### D.3 RP-LI

RP-LI has demonstrated the most diversity in its rules compared to all the other settings we've studied. The simplest rules in this setting take the form:

action = BasicAction(predicates=[lambda x: x == '<source_character>', lambda x: x == '@', lambda x: x == '#'], change_pos=[0], mapping_fn=[lambda x: '<target_character>'])

However, this type of rule comprises only a small portion of the rules in RP-PI. In contrast, most of the rules in IDP-PI are even simpler versions of this rule.

In RP-LI, we observe complex rules like:

action = BasicAction(predicates=[lambda x: x not in ['a', 'e', 'i', 'o', 'u', 'y'], lambda x: x == '@', lambda x: x != 'y'], change_pos=[0], mapping_fn=[lambda x: x + 'y']

These rules not only check for the absence of a character group but also extend beyond this by looking ahead in the sequence and adding a character.

Moreover, there are rules that demonstrate the ability to relate long sequences of up to five characters — some of which might be mapped to a character set in the rule:

action = BasicAction(predicates=[lambda x: x not in ['a', 'e', 'i', 'o', 'u', '@', '#'], lambda x: x == '@', lambda x: x in ['a', 'e', 'i', 'o', 'u', '#']], change_pos=[0], mapping_fn=[lambda x: x + 'n']

Finally, one of the most interesting observations in this setting is the ability to draft rules with multiple change positions:

action = BasicAction(predicates=[lambda x: x == '#', lambda x: x == '@', lambda x: x == 't', lambda x: x == '@', lambda x: x == 'h'], change_pos=[2, 4], mapping_fn=[lambda x: '!', lambda x: '!']

### D.4 IDP V/s RP-RI

In this section, we refer to Table 11, where we compare the rules between two cases at opposite ends of the spectrum discussed in the paper: IDP (representing the substance end) and RP-RI (representing the structure end).

It appears that the rules for RP-RI are able to model more diverse relationships compared to IDP.

In the RP-RI case, we observe the use of many functions to check characters and character sequences. These functions evaluate properties such

as whether a character is a consonant, whether it is empty or valid, whether it is a continuant but not a sonorant, whether it is a velar sound, or whether it is a liquid consonant or not a boundary. This enables RP-RI to derive rules for a wide range of cases. In contrast, IDP rules mainly check for two conditions—whether a character is a consonant or a vowel—and even these checks are less frequent. In RP-RI, however, such checks are present in almost every rule.

Another point of difference, following the trend seen in RP-LI, is that RP-RI can propose changes at multiple positions simultaneously (e.g., change_pos=[0, 2, 4]), while IDP always operates on a single position.

Finally, another interesting observation is that the mapping function in RP-RI appears capable of modeling complex transformations of the source word, exemplified by:

mapping_fn = [lambda c: '!', lambda c: c +'l', lambda c: c + 'k']

# E Experiments

## E.1 Evaluation Metrics

**Edit Distance Reward:** In this section, we will present some additional details about the edit distance reward. The original function proposed by (Luo, 2021) computes the reward for multi-law sound law induction as shown in the equation below, where the initial ancestral forms of the language are denoted by $s^{start}$, the attested form within an intermediate daughter language are $s$, the immediate descendant forms of $s$ are $s^{next}$ (obtained by applying the "action" ($\mathrm{BasicAction}$) or sound change 'a'), and the eventual form is denoted by $s^{end}$:

$$R(s,a) = \frac{dist(s, s^{end}) - dist(s^{next}, s^{end})}{dist(s^{start}, s^{end})}$$

It should be noted that $s, s^{next}, s^{start}, s^{end}$ are all vectors containing all the forms for the language, while the distance function (*dist*) is the aggregated Levenshtein edit distance between all protoforms in the protoform vector $s$ and reflexes in the reflex vector $s'$ in a parent-daughter language pair.

$$dist(s, s') = \sum_{i=1}^{n_v} dist(s_i, s'_i)$$

To derive the simplified version of the reward used in this work we consider a simple step version where the predicted reflexes, are compared against the ground truth reflexes by setting $s = s^{start} = s^{source}$ and $s^{next} = s^{pred}$ for predicted reflexes and $s^{end} = s^{target}$ for the target reflexes.

$$R(s^{source}, s^{pred}, s^{target}) = 1 - \frac{dist(s^{pred}, s^{target})}{dist(s^{source}, s^{target})}$$

| PTK | POC |
|---|---|
| action = BasicAction(<br>predicates=[lambda x: x == 'a'],<br>change_pos=[0],<br>mapping_fn=[lambda x: 'i']) | action = BasicAction(<br>predicates=[lambda x: x == 'a',<br>lambda x: x == '@', lambda x: x == '#'],<br>change_pos=[0],<br>mapping_fn=[lambda x: '\u0259']) |
| action = BasicAction(<br>predicates=[lambda x: x == 'n', lambda x:<br>x == '@'],<br>change_pos=[1],<br>mapping_fn=[lambda x: x+'\u027e']) | action = BasicAction(<br>predicates=[lambda x: x == 'a', lambda x: x<br>== '@', lambda x: x not in ['a','e','i','o','u']],<br>change_pos=[0],<br>mapping_fn=[lambda x: 'e']) |
| action = BasicAction(<br>predicates=[lambda x: x == '\u0294', lambda x:<br>x == '@', lambda x: x == '#'],<br>change_pos=[0],<br>mapping_fn=[lambda x: 'w']) | action = BasicAction(<br>predicates=[lambda x: x in ['a','e','i','o','u'],<br>lambda x: x == '@', lambda x: x == '\u014b',<br>lambda x: x == '@', lambda x: x != '#'],<br>change_pos=[0],<br>mapping_fn=[lambda x: '!']) |
| action = BasicAction(<br>predicates=[lambda x: x in ['a','e','i','o','u',<br>'\u026f','\u0250']],<br>change_pos=[0],<br>mapping_fn=[lambda x: x+'j']) | action = BasicAction(<br>predicates=[lambda x: x not in ['a','e','i','o','u'],<br>lambda x: x == '@', lambda x: x == 'a', lambda x:<br>x == '@', lambda x: x not in ['a','e','i','o','u']],<br>change_pos=[2],<br>mapping_fn=[lambda x: 'u']) |

Table 9: Comparison of generated rules for the RP-PI setting between ptk and poc (The actions are listed in decreasing order of frequency for their type, with the most common occurrences at the top.)

| PTK | POC |
|---|---|
| action = BasicAction(<br>predicates=[lambda x: x == 'i'],<br>change_pos=[0],<br>mapping_fn=[lambda x: '\u0259']) | action = BasicAction(<br>predicates=[lambda x: x == 't'],<br>change_pos=[0],<br>mapping_fn=[lambda x: 'j']) |
| action = BasicAction(<br>predicates=[lambda x: x == 'i'],<br>change_pos=[0],<br>mapping_fn=[lambda x: 'sv']) | action = BasicAction(<br>predicates=[lambda x: x == '\u014b'],<br>change_pos=[0],<br>mapping_fn=[lambda x: '\u0254']) |
| action = BasicAction(<br>predicates=[lambda x: x == 'm'],<br>change_pos=[0],<br>mapping_fn=[lambda x: 'f']) | action = BasicAction(<br>predicates=[lambda x: x == 'u', lambda x:<br>x == '@',<br>lambda x: is_consonant(x)],<br>change_pos=[0],<br>mapping_fn=[lambda x: 'o']) |
| action = BasicAction(<br>predicates=[lambda x: x == 'p'],<br>change_pos=[0],<br>mapping_fn=[lambda x: 'sts']) | action = BasicAction(<br>predicates=[lambda x: is_vowel(x), lambda x:<br>x == '@', lambda x: x == 't'],<br>change_pos=[2],<br>mapping_fn=[lambda x: 'd']) |

Table 10: Comparison of generated rules for the IDP-PI setting between ptk and poc.

| IDP-PI | RP-RI |
|---|---|
| action = BasicAction(<br>predicates=[lambda x: x == 't'],<br>change_pos=[0],<br>mapping_fn=[lambda x: 'j']) | action = BasicAction(\n<br>predicates=[is_consonant],\n<br>change_pos=[0],\n<br>mapping_fn=[lambda c: '!'],\n) |
| action = BasicAction(<br>predicates=[lambda x: x == '\u014b'],<br>change_pos=[0],<br>mapping_fn=[lambda x: '\u0254']) | action = BasicAction(\n<br>predicates=[is_anything, is_nothing, lambda c<br>: c == 'u', is_nothing, is_cont_not_son],\n<br>change_pos=[0, 2, 4],\n<br>mapping_fn=[lambda c: '!', lambda c: '!',<br>lambda c: c+'s'],\n) |
| action = BasicAction(<br>predicates=[lambda x: x == 'u', lambda x:<br>x == '@', lambda x: is_consonant(x)],<br>change_pos=[0],<br>mapping_fn=[lambda x: 'o']) | action = BasicAction(\n<br>predicates=[is_velar, is_nothing,<br>is_liquid_consonant],\n<br>change_pos=[0],\n<br>mapping_fn=[lambda c: c+'b\u02b7'],\n) |
| action = BasicAction(<br>predicates=[lambda x: is_vowel(x),<br>lambda x: x == '@', lambda x: x == 't'],<br>change_pos=[2],<br>mapping_fn=[lambda x: 'd']) | action = BasicAction(\n<br>predicates=[is_not_boundary, is_nothing,<br>lambda c: c == 'p\u02b7', is_nothing, is_son,<br>is_nothing, lambda c: c == 'f'],\n<br>change_pos=[2, 4, 6],\n<br>mapping_fn=[lambda c: '!', lambda c:<br>c+'l', lambda c: c+'k'],\n) |

Table 11: Comparison of diversity in rules between the IDP-PI and RP-RI setting.